\documentclass[letterpaper]{article}
\usepackage[preprint]{aaai2027}
\usepackage[hyphens]{url}
\usepackage{graphicx}
\usepackage{natbib}
\usepackage{caption}
\usepackage{booktabs}
\usepackage{multirow}
\usepackage{amsmath}
\usepackage{amssymb}
\usepackage{xspace}

\title{Reachability Is Not Realization: Tracing the Sources of LLM Benchmark Gains}

\author{
Yanchao Li\textsuperscript{\rm 1,2},
Wanhao Liu\textsuperscript{\rm 2},
Jiaqing Xie\textsuperscript{\rm 2},
Ben Gao\textsuperscript{\rm 2},
Yanbo Wang\textsuperscript{\rm 3},
Tianfan Fu\textsuperscript{\rm 1,2,*},
Yuqiang Li\textsuperscript{\rm 2,*}
}
\affiliations{
\textsuperscript{\rm 1}Nanjing University
\textsuperscript{\rm 2}Shanghai Artificial Intelligence Laboratory\\
\textsuperscript{\rm 3}North University of China\\
\textsuperscript{*}Corresponding authors.
}

\begin{document}
\maketitle

\begin{abstract}
Benchmark gains are often treated as evidence of greater LLM capability.
Yet the same gain can reflect different changes in model behavior.
A model may reach new answers, or produce answers that were already within reach.
Aggregate scores do not distinguish these changes question by question.
We establish a question-level audit under fixed budgets, temperatures, and answer formats.
A question is realized when the default deployment procedure produces the correct answer.
A question is reachable when a specified probe finds that answer within a fixed budget.
We first test whether inference-time layer routing can expand reachability.
Under a matched budget, random routes match or exceed structured search in all 43 model and task settings.
Answer-blind procedures retain almost none of this gain, which instead requires access to the correct answer.
We then ask why reachable answers sometimes fail to appear.
Across six cases spanning 0.5B to 31B, silencing one identified MLP block repairs 68 to 92 percent of a predefined failure set.
We next test whether training closes the gap by expanding reachability.
In five of six matched evaluations, deployed performance rises while the reachable ceiling remains flat or falls.
For DAPO, the deployed score rises by 14.7 points while the reachable ceiling falls by 13.3 points.
Across the settings we audit, realization and reachability therefore do not always change together.
Claims of capability expansion should report both realized performance and reachability under matched evaluation conditions.
Code is available at \url{https://github.com/LiZaiyuan0619/reachability-not-realization}.
\end{abstract}

\section{Introduction}

Benchmark scores remain the main measure of progress in LLM reasoning.
Many recent methods raise these scores without simply increasing model size.
These gains are often interpreted as evidence that a model can solve more questions.
Yet an aggregate score records only how many outputs are correct.
It does not show which questions became newly reachable.
The same gain may instead come from producing already-reachable answers more reliably.
Distinguishing these changes requires following the same questions under matched evaluation conditions.

\begin{figure*}[t]
\centering
\includegraphics[width=0.995\textwidth]{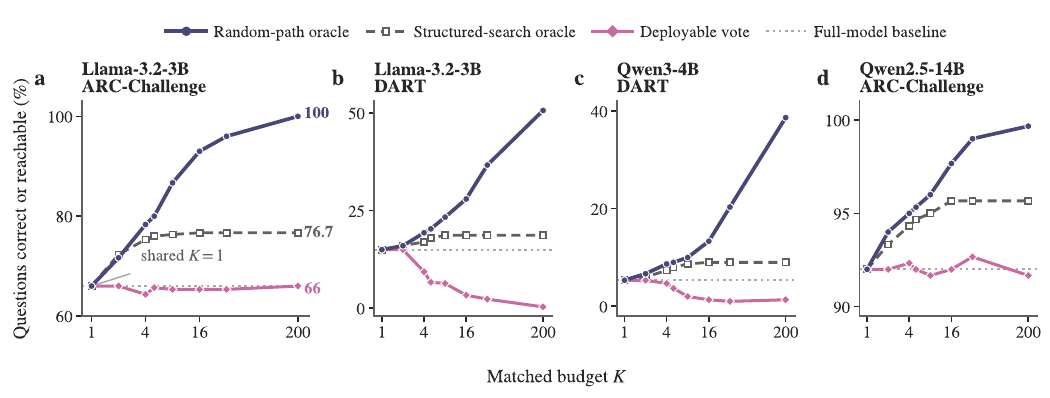}
\caption{\textbf{Budget-matched routing trajectories.}
\textbf{a--d}, Random-path and structured-search oracle reachability,
answer-blind voting, and the full-model baseline across four model--task cells.
All procedures share the same $K=1$ origin. Oracle curves use correctness after
generation and are not deployable selectors.}
\label{fig:routing-budget}
\end{figure*}

We separate these possibilities under fixed budgets, temperatures, and answer formats.
We call a question realized if the default deployment procedure answers it correctly.
We call the same question reachable if a specified probe finds the correct answer within a fixed budget.
Reachability is therefore relative to the evaluation protocol and is not identical to full model capability.
Sampling and alternative layer paths search for correct answers under that protocol.
Recognition provides a separate internal readout and serves only as auxiliary evidence.
This audit asks whether an intervention changes realization, reachability, or both.

Layer routing changes which layers run without changing the weights, providing a controlled inference-time test.
The structured searches we audit are intended to identify useful layer paths.
If their structure matters, they should outperform random routes at the same budget.
We compare structured searches with random routes under the same budget and scoring rule.
Random routes match or exceed the structured ceiling in every evaluated setting.
This comparison is oracle-based because identifying a correct route requires knowing the answer.
The answer-blind procedures we test retain almost none of the oracle gain.
The tested routing gain therefore depends on selecting among reachable outputs.
This explains how the gap can be exploited, but not why it exists.

We next ask why an internal readout can support an answer that direct generation misses.
We predefine questions where the model ranks the correct option first but generates a wrong answer.
In each audited case, silencing one identified MLP block repairs most questions in the predefined failure set.
Silencing attention in the same layer produces a much smaller effect.
Most already-correct questions in a separate healthy panel remain correct under the intervention.
Conversely, perturbing the same block in a healthy model can induce the corresponding failure.
Together, these controls causally localize a direct-answer failure to one MLP block in each case.
The remaining question is whether training closes the broader gap by expanding reachability.

Reinforcement learning with verifiable rewards (RLVR) updates model weights using rewards for correct outputs.
It therefore provides a direct test of whether training expands reachability.
We compare base and trained checkpoints on the same questions under matched answer formats, temperatures, and budgets.
Deployed performance rises in most of these matched evaluations.
The reachable ceiling remains flat or falls in those same evaluations.
Every newly realized question was already reachable in the corresponding base model at the same budget.
The gain also concentrates on questions that the base model reached more frequently.
The audited pairs therefore show that RLVR can improve realization without consistently expanding reachability.

The three analyses share a measurement distinction rather than a common mechanism.
The routing audit tests whether an inference-time gain depends on selection.
The MLP intervention localizes one cause of failed realization.
The RLVR comparison tracks how realization and reachability change with training.
Together, they show that a benchmark gain does not establish that reachability has expanded.
Improving realization remains a meaningful form of progress.
Claims of capability expansion therefore require matched measurements of realized performance and reachability.

We make four contributions.

\begin{enumerate}
\item We establish a question-level audit of realized performance and reachability under fixed conditions.
\item We audit layer routing using budget-matched random routes and answer-blind selection tests.
\item We causally localize a predefined direct-answer failure to one MLP block using component, specificity, and reciprocal controls.
\item We track how realization and reachability change for each question across matched base and RLVR-trained checkpoints.
\end{enumerate}

\begin{figure*}[t]
\centering
\includegraphics[width=0.995\textwidth]{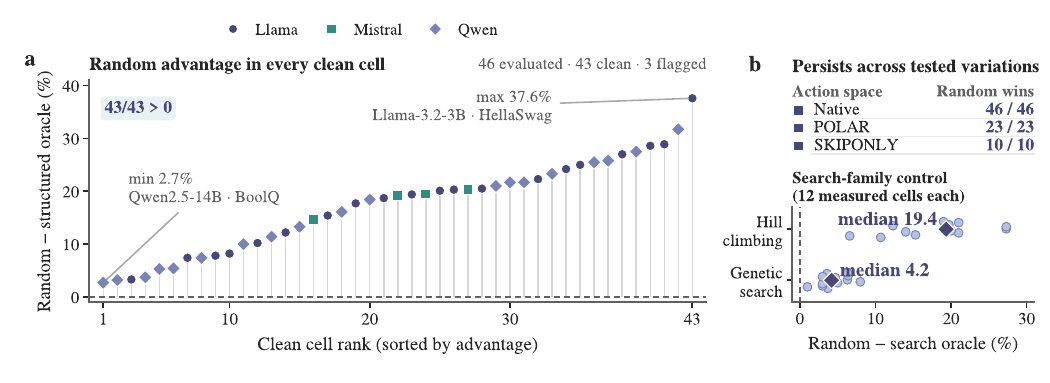}
\caption{\textbf{Breadth and tested robustness of the random-path oracle
advantage.}
\textbf{a}, Random-minus-structured oracle differences across 43 clean
model--task cells; three format-collapse cells are excluded.
\textbf{b}, Directional counts across three action spaces and differences
against hill climbing and genetic search at $K=64$. Dots are cells and bars
are medians.}
\label{fig:routing-breadth}
\end{figure*}

\section{Related Work}

Conditional-compute systems vary transformer depth or token participation to
trade model quality against inference cost.
They learn deployable routers, train residual gates, or verify early exits
\citep{raposo2024mixtureofdepthsdynamicallyallocatingcompute,
elhoushi-etal-2024-layerskip,laitenberger2026what,he-etal-2025-router}.
A narrower line searches for input-specific layer programs or oracle
allocations
\citep{heakl2026drllm,li2026skiplayerloopit,
glavas2024dynamiclayerselectiondecoderonly}.
We compare these searches with budget-matched random paths and measure how much
oracle headroom answer-blind selectors recover.

Candidate-set success and deployed selection answer different questions.
The \textit{pass@k} metric asks whether any candidate succeeds
\citep{chen2021evaluating}.
Without gold correctness, selection can use frequency, semantic modes,
geometric centers, or learned rankings
\citep{wang2023selfconsistency,choi2026modexevaluatorfreebestofnselection,
nguyen2026majorityvotingefficientbestofn,
marina2026boostingselfconsistencyranking}.
Global judge correlation can also overstate item-level candidate recovery
\citep{landesberg2026llmjudgescoreslook}.
Cross-model consensus and trained process verifiers use different access,
supervision, and cost contracts
\citep{liu2026llmsjurycrossmodelconsensus,
kuang2026kvprmefficientprocessreward}.

A separate literature shows that models can contain evidence supporting a
correct answer while producing another response
\citep{park2026bridging,yeom2026hallucinationcommitmentfailurelarger}.
Related causal work patches or steers temporal conflicts, answer-conditioned
features, late rescue, temporal retrieval, reasoning-to-answer flow, or
multi-component effects
\citep{hossain2026rightknowledgewronganswer,
park2026mechelkmechanisticinterpretabilityframework,
deng2026wrongrightlaterescue,huang2026prismeditvectortemporal,
zhang-etal-2025-reasoning,yan2026multicomponentcausaltracinglarge}.
Controlled character counting further links early probe-visible information to
late MLP and attention suppression
\citep{datta2026earlyencodinglatesuppression}.
We instead test a predefined direct-answer failure with component, reciprocal,
specificity, and construction controls.

High-sample evaluations show that low-budget gains can coexist with
high-budget crossover or contraction, and item-level analyses record both
entry and loss
\citep{yue2025does,yuan2026understandingdiversitycollapserlvr}.
The relation changes with the validity criterion, candidate generator,
objective, and training duration
\citep{wen2025reinforcementlearningverifiablerewards,
zhou2026hardjustunreacheddiagnosing,
nguyen2025reasoningboundaryparadoxreinforcement,
yao2025debaterlvrreasoningcapability}.
Teacher-assisted curricula can increase broad-budget coverage, but external
traces change the causal question
\citep{cai2026curriculumreinforcementlearningincentivize}.
We join matched checkpoints by question and track deployment and reachability
under one fixed candidate generator.

\section{Setup}

\paragraph{Evaluation coordinates.}
Reachability is defined relative to a protocol comprising the question set,
answer format, grading rule, temperature, probe, and candidate budget $K$.
Matched comparisons vary only the intervention or checkpoint of interest.
An answer not found by a finite probe is unobserved under that protocol, not
impossible for the model.

\paragraph{Realization and reachability.}
For question $i$, let $g_i(a)\in\{0,1\}$ indicate whether answer $a$ is correct
under the task-specific grading rule.
Let $\hat y_i^{\mathrm{dep}}$ be the output of the designated answer-blind
deployment procedure.
The specified probe returns $K$ candidate outputs, denoted
$c_i^{(1)},\ldots,c_i^{(K)}$.
We define
\[
d_i=g_i\!\left(\hat y_i^{\mathrm{dep}}\right),
\qquad
o_i(K)=\max_{1\leq k\leq K}g_i\!\left(c_i^{(k)}\right).
\]
A question is realized when $d_i=1$ and reachable when $o_i(K)=1$.
The corresponding aggregate quantities are
\[
D=\frac{1}{n}\sum_{i=1}^{n}d_i,
\qquad
O_K=\frac{1}{n}\sum_{i=1}^{n}o_i(K).
\]
We call $D$ deployed performance and $O_K$ oracle reachability, or the
reachable ceiling, at the stated protocol.
The oracle uses correctness only after generation and is diagnostic rather
than deployable.

\paragraph{Question-level comparisons.}
We retain $d_i$ and $o_i(K)$ for every question on both sides of a comparison.
Changes in $d_i$ record newly realized answers and losses, while changes in
$o_i(K)$ record entry into or exit from the measured reachable set.
The oracle--deployment gap is $O_K-D$ when both quantities use compatible
conditions.
An answer-blind selector may use candidates and model-derived scores but not
the correct answer.
Recognition is a separate auxiliary readout that ranks the correct option
under multiple-choice likelihood.
It is neither direct production nor evidence of sampling reachability.

\paragraph{Models and tasks.}
Evaluated checkpoints include Llama 3, Qwen2.5, Gemma 4, Qwen3.5, and OLMo 2
\citep{grattafiori2024llama,qwen2025qwen25technicalreport,team2026gemma,
olmo2024olmo2furious}.
Tasks include ARC-Easy (ARC-E) and ARC-Challenge (ARC-C), GSM8K, PIQA,
LogiQA (LQ), and DART-Math (DART)
\citep{clark2018thinksolvedquestionanswering,
cobbe2021trainingverifierssolvemath,
bisk2019piqareasoningphysicalcommonsense,
liu2020logiqachallengedatasetmachine,
tong2024dartmathdifficultyawarerejectiontuning}.

\paragraph{Matched protocols.}
For layer routing, structured and random paths share the question set, scoring
rule, decode length, initial full path, and candidate budget.
For MLP localization, the failure set is fixed before intervention by correct
recognition and incorrect direct generation.
Component controls, separate healthy panels, and opposite perturbation
directions test localization, selectivity, and reciprocal causality.
For RLVR, base and trained checkpoints are joined by question ID and evaluated
with matched answer formats, temperatures, budgets, and grading rules.
Models, tasks, prompts, run counts, and implementation details are reported in
the Supplement.

\begin{figure*}[t]
\centering
\includegraphics[width=0.995\textwidth]{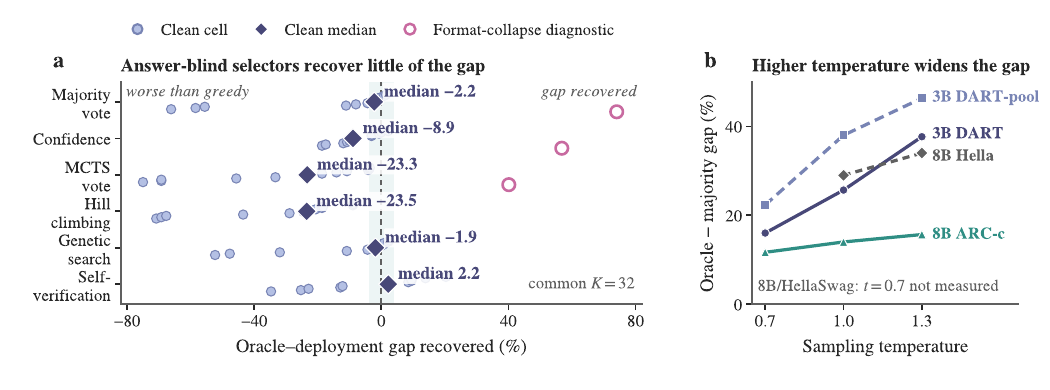}
\caption{\textbf{Answer-blind selectors recover little of the routing oracle
gap.}
\textbf{a}, Recovery of the oracle--greedy gap by six answer-blind channels at
$K=32$, defined as
$(\mathrm{selector}-\mathrm{greedy})/
(\mathrm{oracle}-\mathrm{greedy})$. Bars are clean-cell medians; hollow points
are excluded format-collapse cells.
\textbf{b}, Oracle-minus-majority gaps across measured temperatures.}
\label{fig:routing-selection}
\end{figure*}

\section{Experiment}

\subsection{Routing and Selection}

\noindent\textbf{Random-path reachability continued to rise after the tested
structured search saturated (Figure~\ref{fig:routing-budget}).}
All procedures shared the path budget, scoring rule, decode length, and initial
full path.
Random search provides a necessary control for structured architecture search
\citep{pmlr-v115-li20c}.
The random-path oracle kept improving after the structured search flattened,
reaching 100\% on Llama-3.2-3B/ARC-Challenge at $K=200$.
The answer-blind vote remained near or below the full-model baseline, so
additional correct paths did not provide a deployable selection rule.

\noindent\textbf{The random-path oracle advantage persisted across the
evaluated models, tasks, action spaces, and search families
(Figure~\ref{fig:routing-breadth}).}
The random-path oracle exceeded the structured ceiling in all 43 clean cells
after three format-collapse evaluations were excluded.
The ordering persisted across routing action spaces and against hill climbing
and genetic search.
The effect was therefore not confined to one cell or search design.

\noindent\textbf{Large routing gains appeared only in answer-aware channels
(Table~\ref{tab:reroll}).}
The learned router and answer-blind vote produced little or negative change,
whereas both oracles gained by selecting with correctness.
The table reports channel-specific budgets; the matched comparison is given in
Figures~\ref{fig:routing-budget} and~\ref{fig:routing-breadth}.

\begin{table}[t]
\centering
\setlength{\tabcolsep}{2pt}
\begin{tabular}{@{}lrrrr@{}}
\toprule
channel & ARC-C & ARC-E & DART & pool \\
\midrule
learned router
    & $+0.1$ & $-0.1$ & $-0.1$ & $+0.5$ \\
random vote ($K=16$)
    & $-2.0$ & $\phantom{+}0.0$ & $-8.8$ & $-10.0$ \\
\addlinespace
MCTS oracle (50)
    & $+14.3$ & $+8.7$ & $+6.2$ & $+4.0$ \\
random oracle (200)
    & $+33.7$ & $+15.3$ & $+38.0$ & $+50.0$ \\
\bottomrule
\end{tabular}
\caption{\textbf{Large routing gains appear only with answer-aware selection.}
Percentage-point changes from the unmodified model. The first two rows are
answer-blind; the oracle rows select with correctness. Parentheses give $K$ or
MCTS simulations.}
\label{tab:reroll}
\end{table}

\noindent\textbf{The tested answer-blind selectors recovered little or none of
the oracle--deployment gap (Figure~\ref{fig:routing-selection}a).}
At $K=32$, five of six clean median recoveries were negative; the only positive
median was 2.2\%.
Higher temperature widened the oracle-minus-majority gap
(Figure~\ref{fig:routing-selection}b).
The bottleneck was therefore identifying correct candidates without hindsight.

\begin{table}[t]
\centering
\setlength{\tabcolsep}{3pt}
\begin{tabular}{@{}lrrrr@{}}
\toprule
model & deployed & vote@8 & oracle@8 & gap \\
\midrule
\multicolumn{5}{@{}l}{\emph{GSM8K}} \\
Llama-3.1-8B & 13.3 & 18.7 & 44.7 & 26.0 \\
Llama-3.2-3B & 12.7 & 15.3 & 40.0 & 24.7 \\
Qwen2.5-7B   & 22.7 & 20.7 & 32.0 & 11.3 \\
Qwen2.5-14B  & 32.7 & 32.7 & 35.3 & 2.6 \\
\addlinespace
\multicolumn{5}{@{}l}{\emph{ARC-C}} \\
Llama-3.1-8B & 84.0 & 86.0 & 92.7 & 6.7 \\
Qwen2.5-7B   & 90.7 & 93.3 & 96.0 & 2.7 \\
\bottomrule
\end{tabular}
\caption{\textbf{Ordinary sampling reproduces the selection gap without layer
routing.} All rows use $n=150$, $K=8$, and temperature 0.8. Gap is oracle@8
minus majority@8 in percentage points.}
\label{tab:sampreroll}
\end{table}

\noindent\textbf{Ordinary sampling reproduced the selection gap without layer
routing (Table~\ref{tab:sampreroll}).}
The oracle remained far above majority voting when sampling left substantial
headroom.
The probes therefore found additional correct answers, but the tested
answer-blind procedures did not reliably identify them.
This leaves a different realization failure: a model can support the correct
answer under an auxiliary readout yet fail to produce it directly.

\subsection{MLP Localization}

\begin{figure*}[t]
\centering
\includegraphics[width=0.995\textwidth]{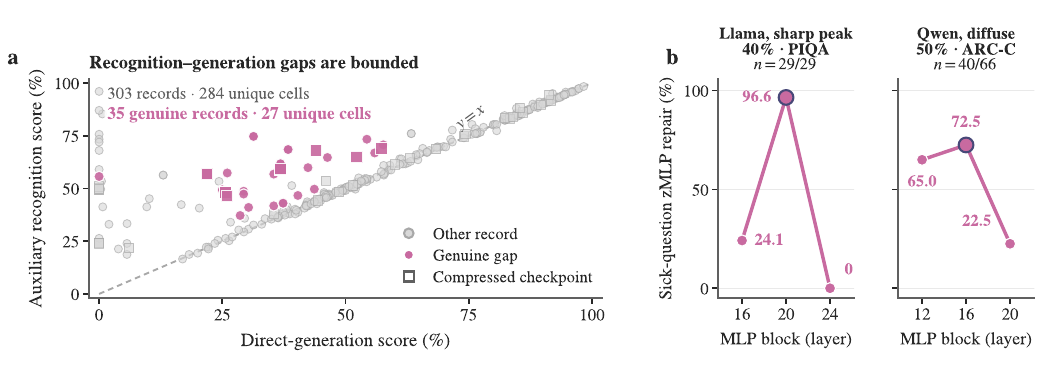}
\caption{\textbf{Bounded occurrence and pruning-created MLP-sensitive
failures.}
\textbf{a}, Recognition versus direct-generation scores across 303 records
from 284 checkpoint--task cells; highlighted records meet the predefined
failure criterion.
\textbf{b}, MLP-silencing repair across layers in two pruning-created cases.
Facet titles give evaluated/eligible counts.}
\label{fig:mlp-occurrence}
\end{figure*}

\noindent Transformer MLP blocks can mediate output-relevant computations in
language models
\citep{geva-etal-2021-transformer,NEURIPS2022_6f1d43d5}.

\noindent\textbf{Recognition--generation failures were bounded but
reproducible in the checkpoint census
(Figure~\ref{fig:mlp-occurrence}a).}
The predefined failure comprised correct recognition and incorrect direct
generation; recognition was not treated as sampling reachability.
Thirty-five of 303 records met this criterion, corresponding to 27 of 284
unique checkpoint--task cells.
The failure was bounded but recurred across checkpoints and tasks.

\noindent\textbf{Pruning created both sharply localized and diffuse
MLP-sensitive failures (Figure~\ref{fig:mlp-occurrence}b).}
Magnitude pruning reproduced the same recognition-correct,
generation-wrong pattern.
Llama-3.1-8B showed a sharp 96.6\% repair peak at layer 20, whereas Qwen2.5-7B
showed a broader response across layers 12 and 16.
The failure was therefore reproducible without implying universal
single-layer concentration.

\begin{figure*}[t]
\centering
\includegraphics[width=0.995\textwidth]{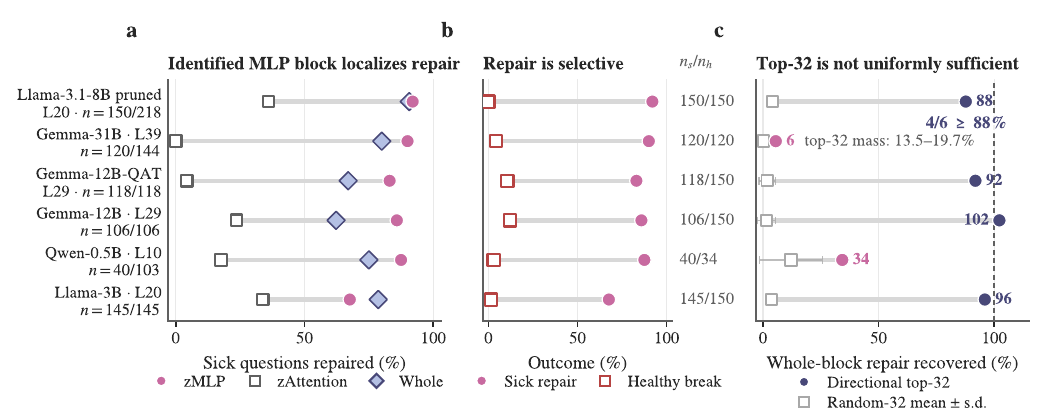}
\caption{\textbf{Selective causal localization to an identified MLP block.}
\textbf{a}, Fraction of the predefined failure set repaired by silencing the
identified MLP block, attention in the same layer, or the whole layer in six
cases.
\textbf{b}, Failure-set repair and healthy-panel breakage under MLP silencing.
\textbf{c}, Directional top-32 and random-32 repair normalized by whole-block
repair; error bars show the random-control standard deviation. Labels give
evaluation counts.}
\label{fig:mlp-localization}
\end{figure*}

\noindent\textbf{Silencing the identified MLP block repaired the predefined
failure more consistently than silencing attention in the same layer
(Figure~\ref{fig:mlp-localization}a; Table~\ref{tab:locus}).}
MLP-block silencing repaired 67.6--92.0\% of the evaluated failure set, and
same-layer attention silencing was lower in all six cases spanning 0.5B to 31B.
Whole-layer silencing was lower in five cases.
These comparisons identify the MLP block as the consistent
component-specific repair locus.

\noindent\textbf{The block intervention was selective for the predefined
failure set and reciprocal across perturbation directions
(Figure~\ref{fig:mlp-localization}b;
Tables~\ref{tab:locus} and~\ref{tab:bidir}).}
MLP silencing broke only 0--12\% of separate healthy panels while repairing
67.6--92.0\% of the failure sets.
Opposite directional perturbations rescued 42.5--97.5\% of failures and induced
10.0--97.5\% failures in healthy panels.
This selectivity, together with reciprocal rescue and induction, supports
case-level causal localization rather than an indiscriminate output shift.

\begin{table*}[t]
\begin{minipage}[t]{0.49\textwidth}
\vspace{0pt}
\centering
\setlength{\tabcolsep}{2pt}
\begin{tabular}{@{}lcrrrr@{}}
\toprule
case
    & \shortstack{$n_s$\\elig./eval.}
    & M & A & L
    & \shortstack{healthy\\$n$/break (\%)} \\
\midrule
\shortstack[l]{G4-31B\\LQ, L39}
    & 144/120 & 90.0 & 0.0  & 80.0 & 120/4.2 \\
\addlinespace
\shortstack[l]{G4-12B\\LQ, L29}
    & 106/106 & 85.9 & 23.6 & 62.3 & 150/12.0 \\
\addlinespace
\shortstack[l]{G4-12B-QAT\\LQ, L29}
    & 118/118 & 83.1 & 4.2  & 67.0 & 150/10.7 \\
\addlinespace
\shortstack[l]{Q2.5-0.5B\\PIQA, L10}
    & 103/40  & 87.5 & 17.5 & 75.0 & 34/2.9 \\
\addlinespace
\shortstack[l]{L3.2-3B\\ARC-C, L20}
    & 145/145 & 67.6 & 33.8 & 78.6 & 150/1.3 \\
\addlinespace
\shortstack[l]{L3.1-8B-pruned\\PIQA, L20}
    & 218/150 & 92.0 & 36.0 & 90.7 & 150/0.0 \\
\bottomrule
\end{tabular}
\caption{\textbf{MLP-block silencing selectively repairs predefined
direct-answer failures.} M/A/L denote MLP, attention, and whole-layer repair
rates (\%). The final column gives healthy-panel size and breakage.}
\label{tab:locus}
\end{minipage}\hfill%
\begin{minipage}[t]{0.49\textwidth}
\vspace{0pt}
\centering
\setlength{\tabcolsep}{2pt}
\begin{tabular}{@{}lccc@{}}
\toprule
case
    & \shortstack{evaluated\\$n_{\mathrm{sick}}/n_{\mathrm{healthy}}$}
    & rescue (\%) & failure (\%) \\
\midrule
\shortstack[l]{G4-12B\\LQ, L29}
    & 40/40 & $95.0\;(-4v)$ & $97.5\;(+4v)$ \\
\addlinespace
\shortstack[l]{G4-12B-QAT\\LQ, L29}
    & 40/40 & $97.5\;(-8v)$ & $95.0\;(+4v)$ \\
\addlinespace
\shortstack[l]{G4-31B\\LQ, L39}
    & 40/40 & $92.5\;(-4v)$ & $27.5\;(+2v)$ \\
\addlinespace
\shortstack[l]{L3.2-3B\\ARC-C, L20}
    & 24/40 & $91.7\;(-8v)$ & $32.5\;(+8v)$ \\
\addlinespace
\shortstack[l]{Q2.5-1.5B\\PIQA, L18}
    & 40/40 & $42.5\;(-4v)$ & $90.0\;(+8v)$ \\
\addlinespace
\shortstack[l]{Q2.5-3B\\PIQA, L28}
    & 20/40 & $70.0\;(-8v)$ & $10.0\;(+8v)$ \\
\bottomrule
\end{tabular}
\caption{\textbf{Opposite directional perturbations provide reciprocal rescue
and induction.} The predefined sick and healthy panels begin at 0\% and 100\%
accuracy. Rescue uses $-\alpha v$ and induction uses $+\alpha v$, with
$\alpha\in\{2,4,8\}$; parentheses give the selected dose.}
\label{tab:bidir}
\end{minipage}
\end{table*}

\noindent\textbf{Directional top-32 units recovered most of the whole-block
effect in four cases, but within-block organization was not uniformly sparse
(Figure~\ref{fig:mlp-localization}c).}
The directional subset recovered at least 87.5\% of whole-block repair in four
of six cases, but only 5.6\% and 34.3\% in the two exceptions.
The selected units carried 13.5--19.7\% of output mass, while random controls
remained small.
The effect can be directionally concentrated, but the data do not support a
universal 32-unit circuit.

\noindent\textbf{The MLP analyses localize this bounded failure without
implying a universal layer or sparse circuit.}
We next move from interventions on fixed checkpoints to weight updates and ask
whether deployed training gains expand reachability or instead change the
realization of answers already within reach.

\subsection{RLVR and Reachability}

\begin{figure*}[t]
\centering
\includegraphics[width=0.995\textwidth]{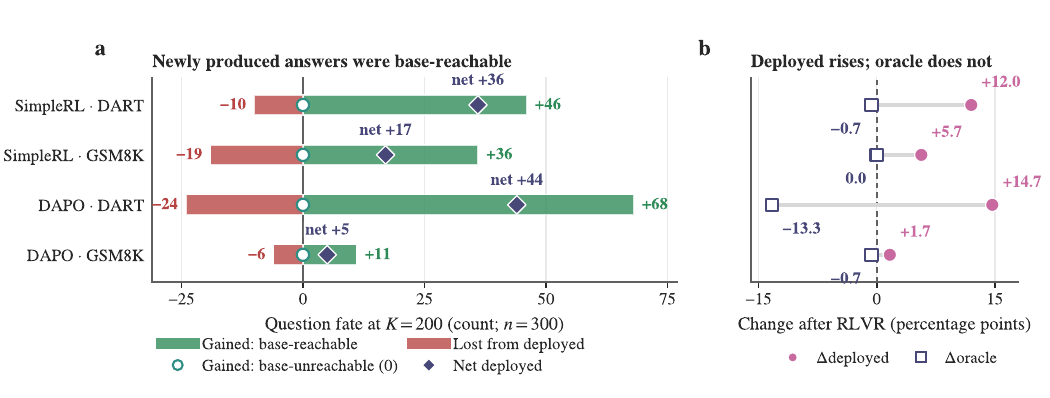}
\caption{\textbf{Question-level fate after RLVR training.}
\textbf{a}, Gains from base-reachable and base-unreachable questions, losses,
and net deployed change.
\textbf{b}, Deployed and oracle changes in four question-matched cells
($n=300$, $K=200$).}
\label{fig:rlvr-fate}

\centering
\setlength{\tabcolsep}{4pt}
\begin{tabular}{@{}llccrrr@{}}
\toprule
matched checkpoint pair & training regime & $n$ & $K$
    & base D/O & trained D/O
    & $\Delta n_{\mathrm{reachable}}$ \\
\midrule
Qwen2.5-32B $\rightarrow$ DAPO-32B
    & from-base RLVR & 300 & 200 & 55.0/96.0 & 69.7/82.7 & $-40$ \\
Qwen2.5-Math-7B $\rightarrow$ SimpleRL-Zero
    & from-base RLVR & 300 & 200 & 68.3/94.0 & 80.3/93.3 & $-2$ \\
Qwen3.5-35B-A3B Base $\rightarrow$ Instruct
    & full post-training control & 150 & 64 & 72.0/83.3 & 68.7/82.7 & $-1$ \\
\bottomrule
\end{tabular}
\captionof{table}{\textbf{Matched checkpoint comparison.}
D/O denotes deployed/oracle accuracy (\%); $\Delta n_{\mathrm{reachable}}$ is
the trained-minus-base reachable count. All rows use matched DART prompts at
temperature 0.8.}
\label{tab:fate}

\medskip
\begin{minipage}[t]{0.48\textwidth}
\vspace{0pt}
\centering
\includegraphics[width=\linewidth]{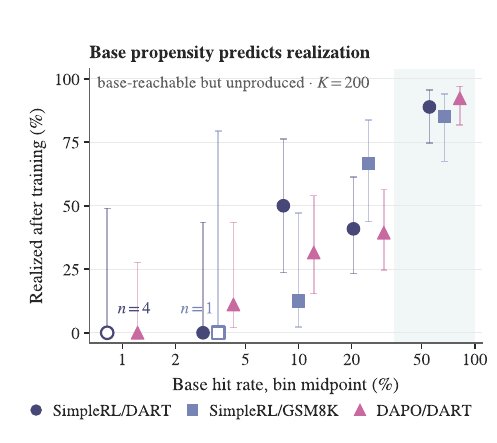}
\captionof{figure}{\textbf{Base propensity predicts post-RLVR realization.}
Realization by base hit-rate bin at $K=200$ with 95\% Wilson intervals.}
\label{fig:rlvr-propensity}
\end{minipage}\hfill%
\begin{minipage}[t]{0.48\textwidth}
\vspace{0pt}
\centering
\setlength{\tabcolsep}{3pt}
\begin{tabular}{@{}lcrrr@{}}
\toprule
pair & $K$ & oracle B/T (\%) & gap (pp) & $\Delta n_r$ \\
\midrule
DAPO & 200 & 95.3/85.3 & $+10.0$ & $-15$ \\
     & 400 & 96.0/85.3 & $+10.7$ & $-16$ \\
\addlinespace
SimpleRL & 200 & 94.0/94.7 & $-0.7$ & $+1$ \\
         & 400 & 94.0/94.7 & $-0.7$ & $+1$ \\
\bottomrule
\end{tabular}
\captionof{table}{\textbf{Sampling-budget control.} Oracle reachability on DART
at $n=150$ and temperature 0.8. B/T denotes base/trained.}
\label{tab:krobust}

\smallskip
\begin{tabular}{@{}llrrr@{}}
\toprule
variant & training & deployed & majority & oracle \\
\midrule
base          & none       & 12.7 & 2.7  & 74.0 \\
RL-zero-math  & RLVR       & 25.3 & 30.0 & 60.7 \\
thinking-SFT  & SFT        & 62.0 & 65.3 & 82.0 \\
full-instruct & SFT$+$RLHF & 67.3 & 66.7 & 85.3 \\
\bottomrule
\end{tabular}
\captionof{table}{\textbf{Training-recipe control.} DART accuracy (\%) for one
OLMo lineage at $n=150$, $K=64$, and temperature 0.8.}
\label{tab:olmorecipe}

\smallskip
\begin{tabular}{@{}lrrrr@{}}
\toprule
task & $n_{\mathrm{eligible}}$ & native (\%) & shared (\%) & retained (\%) \\
\midrule
GSM8K & 125 & 46.4 & 19.2 & 41.4 \\
DART  & 85  & 28.2 & 14.1 & 50.0 \\
\bottomrule
\end{tabular}
\captionof{table}{\textbf{Prompt-format control.} Realization of base-reachable
but initially unproduced questions. Retained is shared/native.}
\label{tab:promptcontrol}
\end{minipage}
\end{figure*}

\noindent\textbf{RLVR gains in the four clean matched cells came from questions
already reachable by the base model (Figure~\ref{fig:rlvr-fate}a).}
The clean analysis joined SimpleRL and DAPO checkpoints on DART and GSM8K at
$n=300$ and $K=200$
\citep{zeng2025simplerlzooinvestigatingtamingzero,yu2026dapo}.
All 11--68 newly produced answers per cell were base-reachable, while 6--24
previously produced answers were lost.
The deployed gain therefore reweighted an existing reachable pool rather than
monotonically accumulating correct answers.

\noindent\textbf{Deployed performance rose without a consistent increase in
the reachable ceiling (Figure~\ref{fig:rlvr-fate}b;
Table~\ref{tab:fate}).}
Deployed performance increased by 1.7--14.7 points across the four cells,
whereas oracle reachability was unchanged or lower.
For DAPO/DART, the two changes were $+14.7$ and $-13.3$ points, corresponding
to 40 fewer reachable questions.
Improved realization occurred in every clean cell; reachability expansion did
not.

\noindent\textbf{A larger sampling budget did not recover the DAPO
reachability loss (Table~\ref{tab:krobust}).}
Doubling $K$ left the trained ceiling unchanged while the base ceiling rose
slightly.
SimpleRL remained stable, so the DAPO contraction was not a low-budget artifact
within this range.

\noindent\textbf{Reachability changed with the training recipe
(Table~\ref{tab:olmorecipe}).}
Math-only RLVR raised deployed performance but lowered oracle reachability.
Both SFT-containing variants raised the ceiling, showing that the response was
conditional on the training recipe.

\noindent\textbf{Prompt format explained part, but not all, of the measured
realization gain (Table~\ref{tab:promptcontrol}).}
A shared base prompt retained 41.4--50.0\% of native-format realization.
Formatting therefore contributed to the gain without accounting for the full
checkpoint difference.

\noindent\textbf{Training most often realized answers that the base model
already sampled frequently (Figure~\ref{fig:rlvr-propensity}).}
Among base-reachable but initially unproduced questions, 85.2--92.3\% in the
$>35\%$ hit-rate bin became produced after training.
Lower-propensity bins generally had lower estimates, although sparse and
intermediate bins were not strictly monotonic.
Training therefore favored answers the base model already sampled frequently
rather than uniformly converting the reachable pool.

\noindent\textbf{Together, the matched fate and propensity results show that
the audited gains changed realization without consistently expanding the
reachable ceiling.}

\section{Conclusion}

We separated answers that a model produced from those reached by a specified
probe under matched conditions.
Layer routing exposed a selection gap, and causal interventions localized one
bounded realization failure to an MLP block.
Across four clean RLVR comparisons, newly realized answers were already
reachable in the corresponding base model, while the reachable ceiling
remained flat or fell.
Improved realization is meaningful, but reporting should distinguish it from
reachability expansion.

\clearpage
\section*{Acknowledgments}
We thank Zhehong Ai for helpful discussions.

\bibliography{ref}

\end{document}